\documentclass{comjnl}

\usepackage{amsmath}
\usepackage{mathrsfs} 
\usepackage{amssymb}
\usepackage{booktabs}
\usepackage{multirow}
\usepackage[noend]{algpseudocode}
\usepackage{algorithmicx,algorithm}
\usepackage{color}
\usepackage[numbers,sort&compress]{natbib}

\begin{document}

\title[Article Title]{2L-LSH: A Locality Sensitive Hash function-based Method for Rapid Point Cloud Indexing}

\author{Shurui Wang}
\author{Yuhe Zhang \thanks{Corresponding author: zhangyuhe0601@nwu.edu.cn}}
\author{Ruizhe Guo}
\author{Yaning Zhang}
\author{Yifei Xie}
\author{Xinyu Zhou \thanks{Corresponding author: xinyuzhou77@163.com}}
\affiliation{School of Information Science and Technology, Northwest University, Xi’an, PR China} 
\email{zhangyuhe0601@nwu.edu.cn, xinyuzhou77@163.com}

\shortauthors{S.R. Wang}

\received{00 January 2009}
\revised{00 Month 2009}

\keywords{Locality sensitive hash; Point cloud; 3D data indexing; {\it k}-nearest neighbors; Radius neighbors}

\begin{abstract}
The development of 3D scanning technology has enabled the acquisition of massive point cloud models with diverse structures and large scales, thereby presenting significant challenges in point cloud processing. Fast neighboring points search is one of the most common problems, which is frequently used in model reconstruction, classification, retrieval and feature visualization. Hash function is well-known for its high-speed and accurate performance in searching high-dimensional data, which is also the core of the proposed 2L-LSH. Specifically, the 2L-LSH algorithm adopts a two-step hash function strategy, in which the popular step divides the bounding box of the point cloud model and the second step constructs a generalized table-based data structure. The proposed 2L-LSH offers a highly efficient and accurate solution for fast neighboring points search in large-scale 3D point cloud models, making it a promising technique for various applications in the field. The proposed algorithm is compared with the well-known methods including Kd-tree and Octree, the obtained results demonstrated that the proposed method outperforms Kd-tree and Octree in terms of speed, i.e. the time consumption of {\it k}NN search can be 51.111\% and 94.159\% lower than Kd-tree and Octree, respectively. And the RN search time can be 54.519\% and 41.840\% lower than Kd-tree and Octree, respectively.
\end{abstract}

\maketitle

\section{Introduction}\label{sec1}

The point cloud model is a set of discrete 3D points, which usually does not contain topology information. Compared with the triangular meshes, the point cloud model has the advantages of simplicity, flexibility and high precision. In recent years, the rapid development of 3D scanners has gained a great interest in the studies based on point cloud data such as autonomous driving \cite{bib1}, robotics \cite{bib2} and object segmentation \cite{bib3}, \textit{etc}.

 The processing of point cloud usually involves point cloud simplification \cite{bib29}, denoising \cite{bib30}, registration \cite{bib28},  recognition \cite{bib4} and retrieval \cite{bib3}, {\it etc}. 
 
 However, when dealing with the above tasks, it is necessary to search the neighboring points, for example, for calculating local geometric features such as normal and curvature, which is the basis of several processing. The calculation speed and accuracy of neighboring points searching can significantly affect the performance of point cloud processing. However, the lack of topological information in discrete point cloud data makes point cloud neighboring searches very time-consuming. Therefore, an efficient point cloud data storage structure is of great significance for efficient neighboring search\cite{bib5,bib6,bib7}.

 Most of the existing methods use the tree structure-based methods to perform the indexing of points, such as Octree \cite{bib5} and Kd-tree \cite{bib6}. However, the layer of the tree considerably affects the performance of speed, accuracy and memory cost, making it difficult to simultaneously balance these performances. An alternative is to index the point data using hash functions \cite{bib3, bib7}. Among them, locality sensitive hash (LSH) and its variants are considered to be the state-of-the-art data search algorithm for high-dimensional data. In recent years, LSH-based methods showed an admirable performance in both 2D \cite{bib9} and 3D \cite{bib10} spatial data indexes. However, there still exists some drawbacks, such as the large complexity, high computational load of the mapping process, high memory cost, as well as the lack of efficient methods aiming at point cloud models. This is because choosing which kind of hash functions considerably affect the time and space complexity of the algorithm. Furthermore, due to the structural diversity of real-world data, it is always difficult to find a hash function suitable for all kinds of structures.

LSH aims at designing appropriate hash functions to map the objects in order to create more collision conflicts for those with high similarity, and less for those with low similarity \cite{bib4}. Considering the point cloud model, when establishing the hash table, all points in the model are hashed through the hash function and then stored in the corresponding hash bucket. For search operations, the same hash function is used to map the search point to obtain the candidate set, which consists of points in the hash bucket corresponding to the search point. The search result is then achieved by calculating the distance between the search point and each of the candidate points. The advantages of this method reside in the fact that it can greatly reduce the search space, thereby improving the search efficiency.

Based on the above analysis, in this paper, a two-level locality sensitive hash function (2L-LSH) method, which designs a hash function based on the oriented bounding box, is proposed. In the Each small level, we divide the OBB of the model into 24 blocks, and each block is a quadrangular pyramid composed of the connecting line between the coordinate axes, the coordinate origin point, vertexes of the oriented bounding box (OBB) and its projection points on the coordinate axes. Under this division strategy, the volume of the subspace in the corner is smaller and can greatly reduce the waste of time caused by a large number of searches in the blank space. Then, the second-level slices the blocks created in the first level and projects each point into a bin according to its coordinate information. The slice direction parallels the coordinate plane, creating a regular neighboring relationship that speeds up the search for adjacent points.

The main contributions of this paper can be summarized as follows:
\begin{itemize}
        \item  
          A simple and easy-to-implement method 2L-LSH is proposed for fast neighboring search in point clouds, which only takes point coordinates as input.
        \item
             {\it k}NN and RN searches are designed and implemented based on the proposed 2L-LSH.

        \item 
             Compared to classical algorithms, 2L-LSH has advantages such as low memory consumption, fast neighboring search, and performing well on point clouds with different structures and scales.
\end{itemize}

\section{Related works}\label{sec2}

By indexing points in the point cloud model, the speed of point search can be greatly accelerated. The methods of model indexing are mainly divided into two categories: tree structure-based methods and hash function-based methods.

\subsection{Tree structure-based methods}\label{sec2_1}

Among the tree-based point cloud data structure, Octree\cite{bib5} and Kd-tree\cite{bib6} are the most commonly used. The tree data structure divides point cloud data into hierarchical units, thereby accelerating the neighboring points searches.

 Octree is a structure that iteratively divides the three-dimensional space and builds an index. Except for leaf nodes, each node contains eight sub nodes. The establishment principle of Octree is to cut every dimension of (sub)space in half until the predefined tree height is reached or the number of points inside the leaf node is less than the given value. This method stands out due to its ease of implementation and strong operability. When the point cloud data space is evenly distributed and the amount of data is not large, Octree can obtain better search performance. However, when the spatial distribution of point cloud data is uneven, the generated Octree structure will be unbalanced, and it is difficult to execute searches or other operations effectively. In addition, when dealing with a large volume of data, defining numerous tree layers may lead to longer search paths, thereby reducing efficiency. Moreover, the memory occupation of Octree will increase exponentially with the increase of tree layers. If a small number of tree layers is defined, it will result in a large number of points in the leaf nodes, making the point search more time-consuming. 

 Kd-tree is a binary tree of {\it K}-dimensional spatial point data, which is a structure extending from Binary Space Partitioning Trees (BSP Trees). Kd-tree divides the target space into smaller areas with approximately equal amounts of data, thereby establishing neighborhood relationships within the point cloud. However, when dealing with massive data, the computational cost of recursively establishing neighborhood relationships can be really high, and the increase of depth may lead to the reduction of efficiency, which affects the overall performance of the algorithm.

In the neighboring points searching, Octree and Kd-tree have their own advantages and disadvantages. Kd-tree inherits the high search efficiency feature of a binary tree and excels in point search and neighboring points search speed. Compared with Kd-tree, the advantage of Octree is its ability to terminate the search in advance. After searching down to the leaf node, Kd-tree needs to perform a backtracking operation to judge whether the search has been completed, preventing the omission of eligible points. In contrast, Octree can construct a sphere with the search point as the center and the worst-case distance as the radius. If the sphere completely falls into a cube, it can be considered that the search scope has been restricted within the cube. There are also studies on integrating these two types of tree structures to amplify their respective advantages to a certain extent and improve the overall performance. However, both Octree and Kd-tree are prone to the problem of excessive depth and reduced efficiency when facing massive data. After integrating the two types of tree structures, it is still hard to avoid this problem.

\subsection{Hash function-based methods}\label{subsec2_2}

There are several types of model indexing methods based on hash function. Although several studies have been performed on images \cite{bib11} and videos \cite{bib12}, few studies on 3D models exist. 

The method proposed by Tarmissi et al. \cite{bib7} consists in decomposing the Laplace matrix of each sub-mesh of the three-dimensional triangular mesh, in order to calculate its hash value. Lee et al. \cite{bib13} propose a method to generate binary codes with uniqueness and security for the 3D mesh model, which is more robust to mesh attacks. Both the methods proposed in \cite{bib7} and \cite{bib13} are designed for 3D mesh data. However, for point cloud data, they need to be transformed into mesh models before using these methods for data indexing. This process increases the operating time and complicates the algorithm. 

Shao et al. \cite{bib3} propose a framework which combines Perfect Spatial Hash (PSH) with Convolutional Neural Network. This method is suitable for the comparison between models in order to develop functions such as model classification. In \cite{bib9}, a binary code is generated for 2D or 3D data, and the Hamming distance is used to judge the similarity of the model. However, generating binary codes for non-binary data can be challenging. 

There are also various types of data indexing methods based on hash function, with the LSH \cite{bib14} algorithm being the most commonly used in practical contexts. The method proposed by Matei et al. \cite{bib4} can be used for real LiDAR(Light Laser Detection and Ranging) data, where LSH is applied in the feature space in order to perform the Approximate Nearest Neighbor (ANN) search of the descriptors. It has a high accuracy and speed in 3D recognition. Dasgupta et al. \cite{bib15} propose a fast way to estimate the Euclidean distance between two d-dimensional vectors, so as to speed up the Euclidean realization of LSH. Christiani \cite{bib16} compares the Dahlgaard-Knudsen-Thorup framework \cite{bib17} with the Indyk-Motwani framework \cite{bib14}\cite{bib18}, and proves that the former has a better performance, especially when locality-sensitive hash functions are expensive to evaluate. Zheng et al. \cite{bib10} combine LSH with tree structure, where the efficient ANN search is performed by establishing tree structure in the projection space generated by LSH. However, these methods all face the same problem. That is, in order to ensure that the algorithm has a satisfactory performance in speed and accuracy, the memory occupation of its hash table can be very large. Pan et al. \cite{bib19} combined RP-tree with LSH and proposed a bi-level data structure for {\it k}NN search of images. The work of He et al. \cite{bib20} is aimed at the point cloud management system, building the data structure based on Octree and skip list, and designing the corresponding search algorithm to search the nearest neighbors of the current model in the whole dataset. Although the target issue of \cite{bib19} and \cite{bib20} is not the neighboring points searching in three-dimensional point cloud models, their idea of combining the tree structure with the linear structure gives great inspiration to our work.

\section{The proposed LSH scheme}\label{sec3}

\subsection{Overview}\label{sec3_1}
In order to improve the performance of neighboring points searching on 3D point cloud, 2L-LSH, an easy-implemented and low-cost hash function-based data structure, is proposed. In this framework, the oriented bounding box (OBB) of the point cloud is first computed, then a local reference frame (LRF) according to the spatial information of the OBB is built, and the subsequent two-level hash function process is completed based on this LRF.

\subsection{Oriented bounding boxes generation}\label{sec3_2}
\begin{figure}
	\centering
	\includegraphics[width=3.3in]{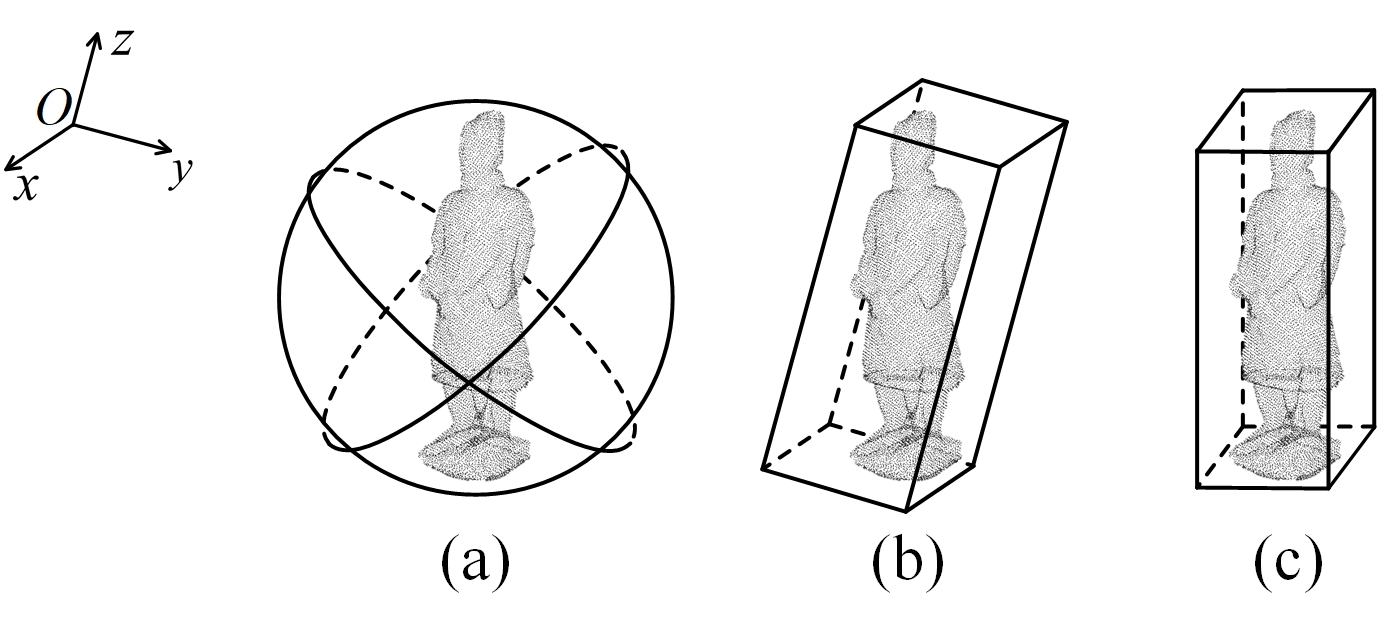}
	\caption{Commonly used bounding volumes. (a) Bounding sphere. (b) AABB. (c) OBB.}\label{fig2}
\end{figure}

The purpose of constructing a bounding volume for the model is to approximate the complex point cloud model to a simple geometric shape, so as to reduce the difficulty of 3D virtual division. Commonly used bounding volume types include bounding spheres \cite{bib21}, axis aligned bounding boxes (AABB) \cite{bib22} and Oriented bounding boxes (OBB) \cite{bib22}. The bounding sphere uses a sphere to enclose the object, while AABB and OBB use rectangular prisms (cf. Figure~\ref{fig2}). It can be seen that although the bounding sphere and AABB can be easily built, they both contain too much “empty space”. This bad approximation of the contained object may lead to problems such as memory waste and search performance degradation in the subsequent data structure design. In contrast to the AABB, whose edges are parallel to the coordinate axes $x$, $y$ and $z$, the edges of the OBB can have an arbitrary length and orientation. As shown in Figure~\ref{fig2}c, by selecting the appropriate orientation, OBB can obtain a better approximation of the object with a minimal volume, and has good adaptability to targets of various shapes.

The key to build the OBB is the selection of the appropriate orientation and the determination of the smallest bounding box in this orientation. This paper applies the PCA \cite{bib22} on the 3D point cloud model.

For an input point cloud model $M \in \mathbb{R}^{m \times n}$, where {\it m} is the number of points contained therein and $n$ is the dimension of each point. A PCA-based bounding box calculation is generally solved by covariance matrix. However, due to the fact that the calculation scale of the covariance matrix will sharply increase with the increase of the data size, the Singular Value Decomposition (SVD) is used rather than the covariance method \cite{bib23}.

The center of the OBB is denoted by point $p_{center}$, and the vertex of the OBB of the given point cloud is denoted by $Ver=\{v_i(x_i,y_i,z_i)\},(i=1,2,\cdots,8)$, and while both are based on the original coordinate system $O_{xyz}$ of the point cloud model. 

\subsection{LRF construction}\label{sec3_3}

\begin{figure}
	\centering
	\includegraphics[width=3.3in]{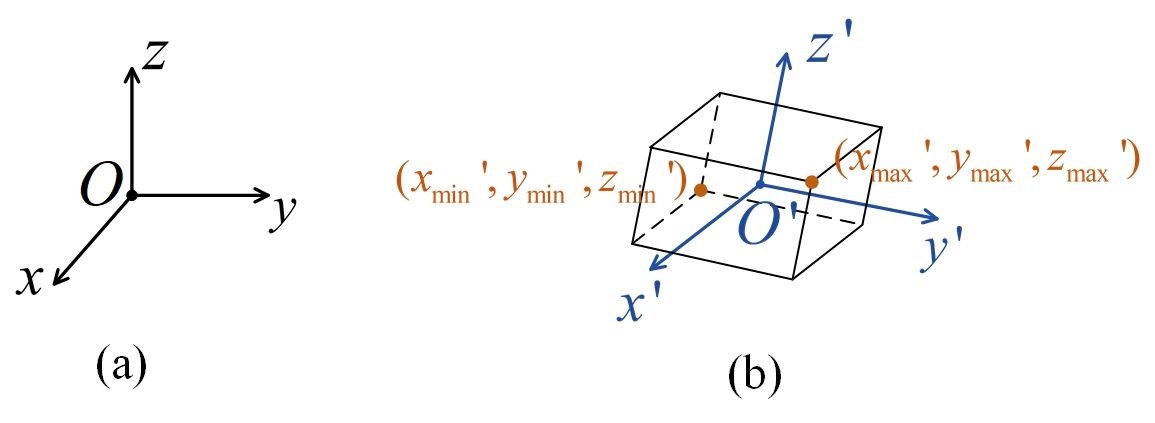}
	\caption{The original (a) and projection (b) coordinate system.}
	\label{fig3}
\end{figure}

Given the OBB, the new coordinate, i.e., the LRF which will be used in the next steps, is created. More precisely, the center point $p_{center}$ of the OBB is considered as the origin  $O^{\prime}$ of the LRF, and three orthogonal edges parallel to the OBB bounding box are considered as the axes {\it x}, {\it y} and {\it z}, thus forming the LRF ${O_{xyz}}^{\prime}$, as shown in Figure~\ref{fig3}, where $({x_{max}}^{\prime}, {y_{max}}^{\prime}, {z_{max}}^{\prime})$ and $({x_{min}}^{\prime}, {y_{min}}^{\prime}, {z_{min}}^{\prime})$ are the diagonal vertices of the OBB.

\subsection{2L-LSH}\label{sec3_4}
\begin{figure}
	\centering
	\includegraphics[width=3.3in]{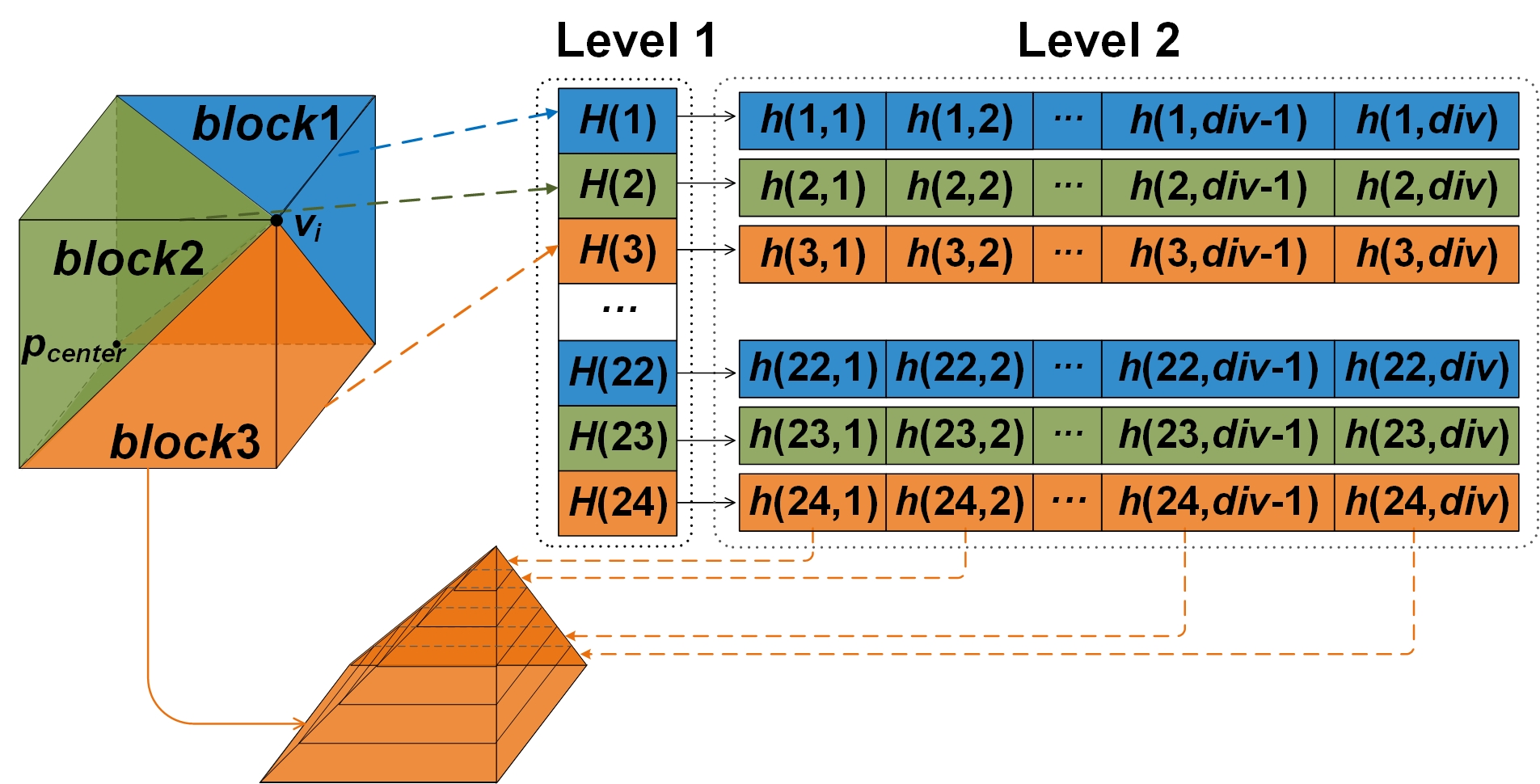}
	\caption{
            General view of hash tables of 2L-LSH.
        }\label{fig1}
\end{figure}

Given the LRF ${O_{xyz}}^{\prime}$, which is constructed in the previous step, this section details the process for building 2L-LSH. The overall structure of the hash table is illustrated in Figure~\ref{fig1}. This process makes the points close to each other falling into the same or adjacent bin. Consequently, only a small range of bins needs to be searched during the neighboring points searching operation, promoting the neighboring points searching efficiency.

\begin{figure}
	\centering
	\includegraphics[width=3.3in]{fig4.jpg}
	\caption{Schematic diagram of 3D data structure construction based on 2L-LSH. (a) 8 quadrants of the projection coordinate system. (b) Divide each quadrant into 3 blocks. (c) Divide each block into $div$ bins(Take $div=5$ as an example).}
	\label{fig4}
\end{figure}

2L-LSH is a two-level hash table. The construction process contains two main steps. In the first step, the first-level hash table $\mathscr{H}=\{H(i)\}, (i=1,2,\cdots,24)$ is generated. More precisely, the LRF ${O_{xyz}}^{\prime}$ divides the OBB of the model into 8 quadrants (cf. Figure~\ref{fig4}a). Each quadrant is then divided into three pyramid shaped blocks that intersect at $p_{center}$, where the corresponding quadrant cube is denoted by $v_i$ (cf. Figure~\ref{fig4}b). Thus, all points in the point cloud can be mapped into hash table $H(i)$. The purpose of this step is to establish a mapping space that best conforms to the shape of the target model, so that it will not occupy too much storage space. The calculation process is summarized as follows.

\begin{equation}
	p\_block=3 \cdot p\_qua+b \label{equ2}
\end{equation}

\begin{equation}
	b=\left\{
		\begin{split}
		1,&\text{  if } -\frac{{y_{p\_qua}}^{\prime}}{{x_{p\_qua}}^{\prime}}\cdot {{x_p}^{\prime}}+{{y_p}^{\prime}}>0\text{ and } \\
		&-\frac{{z_{p\_qua}}^{\prime}}{{x_{p\_qua}}^{\prime}}\cdot {{x_p}^{\prime}}+{{z_p}^{\prime}}>0 \\
		2,&\text{  if } -\frac{{y_{p\_qua}}^{\prime}}{{x_{p\_qua}}^{\prime}}\cdot {{x_p}^{\prime}}+{{y_p}^{\prime}}\leq0\text{ and } \\
		&-\frac{{z_{p\_qua}}^{\prime}}{{y_{p\_qua}}^{\prime}}\cdot {{y_p}^{\prime}}+{{z_p}^{\prime}}>0 \\
		3, &\text{  otherwise} 
		\end{split}
		\right. \label{equ3}
\end{equation}
For each point $p(x_p, y_p, z_p)$ on the point cloud {\it P}, its coordinate $p^{\prime}({x_p}^{\prime}, {y_p}^{\prime}, {z_p}^{\prime})$ in the LRF ${O_{xyz}}^{\prime}$ is calculated by $p^{\prime}=pV$ , where {\it V} is the right rotation matrix generated in the process of calculating OBB. As shown in Figure~\ref{fig4}a, point $p^{\prime}$  locates in one of the 8 quadrants of ${O_{xyz}}^{\prime}$. Denote the quadrant where $p^{\prime}$ locates in by $p\_qua \in \{1,2,\cdots,8\}$, and denote the vertex of OBB corresponding to quadrant $p\_qua$ by $v_{qua}(x_{p\_qua}, y_{p\_qua}, z_{p\_qua})$, the corresponding block $p\_block$ of $p^{\prime}$ in the hash table $\mathscr{H}$ is then calculated using Equation(\ref {equ2}) and (\ref{equ3}).

Given $H(i)$ of point $p^{\prime}$ (cf. Figure~\ref{fig1}), the second-level hash table $H(i)=h(i,j), \quad (i=1,2,\cdots,24;\quad j=1,2,\cdots,div)$ is then generated.  Each block is first divided into {\it div} bins along the edges of the quadrilateral perpendicular to the bottom, as shown in Figure~\ref{fig4}c. The purpose of this step is to limit the number of points in the bin, so as to accelerate the search speed of the neighboring points. All the bins in a block are stored in the same first-level hash table $H(p\_block)$ , while the detailed information of points contained in each bin is stored in the same second-level hash table $h(p\_block, p\_proj)$. That is to say, point $p$ finally falls into the bin indexed by $(p\_block, p\_proj)$. This is to find a suitable division interval, so that the feature of location similarity can be maximized in subsequent operations and the search scope can be reduced as much as possible. Equation (\ref{equ4}) is used to calculate $p\_proj$:
\begin{equation}
	p\_proj=\left\{\begin{array}{l}
		{div}-\lfloor{\frac{\lvert x_p{ }^{\prime}\rvert \cdot div}{len\_x}} \rfloor, \text {  if } b=1 \\
		{div}-\lfloor{\frac{\lvert y_p{ }^{\prime}\rvert \cdot div}{len\_y}} \rfloor, \text {  if } b=2 \\
		{div}-\lfloor{\frac{\lvert z_p{ }^{\prime}\rvert \cdot div}{len\_z}} \rfloor, \text {  if } b=3
	\end{array}\right. \label{equ4}
\end{equation}
where $len\_x$, $len\_y$ and $len\_z$ are 0.5 times the side length of OBB.

As shown in Figure~\ref{fig4}c, we divided each block into $div$ bins along the edges of the quadrilateral which is perpendicular to the bottom in the second level of 2L-LSH (in Section~\ref{sec3_4}). Each bin at the corner of the OBB is a four prism shaped space containing 5 planes (cf. Figure~\ref{fig12b}a), whereas each of the remaining bins is a hexahedral space containing 6 planes (cf. Figure~\ref{fig12b}b). 

\subsection{\textbf{{\it k}}NN and RN search strategy}\label{sec3_5}
The original point cloud data is a collection of a series of discrete points without topological information. Several operations based on the point cloud such as normal estimation, mollification and noise removal, require to process lot of {\it k}NN \cite{bib6} searches, which is usually defined as finding several targets based on nearest distance from the search point. Therefore, an efficient {\it k}NN search strategy is crucial for point cloud processing. 

\begin{figure}
	\centering
	\includegraphics[width=2in]{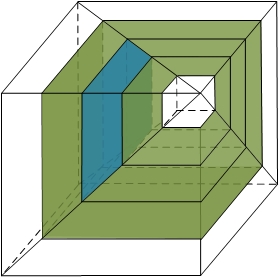}
	\caption{Neighboring bins (marked green) of the bin marked blue.}
	\label{fig5}
\end{figure}

In {\it k}NN and RN search, the 2L-LSH presented in Section~\ref{sec3_4} should be first used to calculate the hash table information where the search point $q(x,y,z)$ is located. Denote the block and bin where point $q$ is hashed into by $q\_block$ and $q\_proj$, respectively, which means that neighboring points of point $q$ have a high probability of locatng in the neighboring bins of bin $(q\_block, q\_proj)$. As shown in Figure~\ref{fig5}, the blue part indicates the bin where point $q$ falls into, and the green part indicates its neighboring bins. The bin $(q\_block, q\_proj)$ and its neighboring bins will be searched according to Algorithm~\ref{algo1} and Algorithm~\ref{algo2}, in which $min\_dist\_boun$ represents the minimum distance between the search point and all planes of the bin currently being searched.

$min\_dist\_boun$ can be calculated as follows. Use $Planes=\{ pla_j \}$ represents the planes of the bin that is being searched, where $j=1,2,\cdots,5$ for bins at the corner of the OBB and $j=1,2,\cdots,6$ for the others. Denote a point in plane $pla_j$ by $p_{pla}$ and the normal vector of plane $pla_j$ by $\overrightarrow{n}$. The distance $dist_i$ from search point $q$ to the plane $pla_j$ can be calculated by Equation (\ref{equ8}). The Euclidean distance set composed of the distance between the search point $q$ and all planes $pla_j$ of the bins that have been searched is denoted by $Dist=\{ dist_i \}$. And then the $min\_dist\_boun$ can be achieved by $min\_dist\_boun$=MIN($Dist$).

\begin{equation}
    dist_i = \frac
            {\lvert \overrightarrow{qp_{pla}} \cdot \overrightarrow{n} \rvert }
            {\|  \overrightarrow{n} \| }
    \label{equ8}
\end{equation}

\begin{figure}
	\centering
	\includegraphics[width=3.3in]{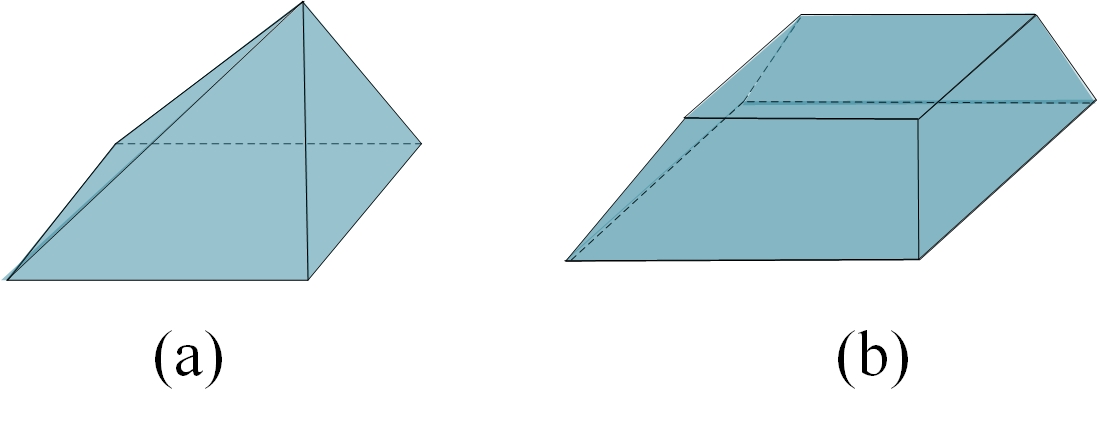}
	\caption{Schematic diagram of the shape of the bin at (a) and not at (b) the corner of the OBB.}
	\label{fig12b}
\end{figure}

\textbf{\textit{k}NN Search}: In {\it k}NN search, search the bin $(q\_block, q\_proj)$ and add the qualified points to the result list $list\_knn$ at first. Then calculate the $min\_dist\_boun$ with $q$ for each neighboring bins of $(q\_block, q\_proj)$. If $min\_dist\_boun$ is less than the distance between $q$ and the {\it k}-th nearest neighboring point currently found, or the number of nearest neighboring points currently found is less than $k$, add this neighboring bin to the search list $list\_search$. Afterwards, the same operation is repeated until $list\_search$ is empty. The specific implementation process of the {\it k}NN search is provided in Algorithm~\ref{algo1}. 

\begin{algorithm}
	\caption{{\it k}NN search}\label{algo1}
	\begin{algorithmic}[1]
		\Require Hash table $\mathscr{H}$, expected number $k$, search point $q$
		\Ensure search result $list\_knn$ 
		\State get the bin indexed by ($q\_block$, $q\_proj$) where $q$ falls into
		\State add ($q\_block$, $q\_proj$) to $list\_search$;
		\While{$list\_search$ is not empty}
		\State $ (blo, pro) \Leftarrow list\_search.pop $
		\State Set {\it list} consists of all neighboring bins of ({\it blo}, {\it pro})
		\For{each ($l\_b$, $l\_p$) in $list$}
		\State $ min\_dist\_boun$ $\Leftarrow $ Shortest distance whitin bin $ (l\_b, l\_p) $
		\State $ dist\_k$ $ \Leftarrow$ Distance of the {\it k}-th nearest neighbor currently found
		\If{$list\_knn.size<k$ or $min\_dist\_boun < dist\_k$}
		\For{each point $p\_temp$ in bin ($l\_b, l\_p$)}
		\If{$dist(p\_temp, q) < dist\_k$ or $list\_knn.size<k$}
		\State update $list\_knn$ with $p\_temp$;
		\EndIf
		\EndFor
		\State update $list\_search$ with $(l\_b, l\_p)$;
		\EndIf
		\EndFor
		\EndWhile
		\State \Return{$list\_knn$}
	\end{algorithmic}
\end{algorithm}

\textbf{RN search}: The RN search consists in finding all the points within radius $r$ for a given point $q(x,y,z)$, which is an indispensable operation in several algorithms using three-dimensional data \cite{bib24}. The concepts of RN search and {\it k}NN search are roughly the same. The difference is that the condition for judging whether to add a bin to the search list. Simpler than {\it k}NN search, RN search only needs to judge whether the distance between the current bin and $q$ is less than radius $r$. Specifically, first hash search point $q$ into bin $(q\_block, q\_proj)$ and scan this bin and add the point whose distance from point $q$ is less than $r$ to the result list $list\_rn$. Then for each neighboring bin of $(q\_block, q\_proj)$, calculate its $min\_dist\_boun$ with $q$. If $min\_dist\_boun$ is less than $r$, add this neighboring bin to the search list $list\_search$. Otherwise, skip to the next bin. Repeat the above process until $list\_search$ is empty. The RN search strategy is presented in Algorithm~\ref{algo2}. 

\begin{algorithm}
	\caption{RN search}\label{algo2}
	\begin{algorithmic}[1]
		\Require Hash table $\mathscr{H}$, expected number $r$, search point $q$
		\Ensure search result $list\_rn$ 
		\State get the bin indexed by ($q\_block$, $q\_proj$) where $q$ falls into
		\State add ($q\_block$, $q\_proj$) to $list\_search$;
		\While{$list\_search$ is not empty}
		\State $ (blo, pro) \Leftarrow list\_search.pop $
		\State Set {\it list} consists of all neighboring bins of ({\it blo}, {\it pro})
		\For{each ($l\_b$, $l\_p$) in $list$}
		\State $ min\_dist\_boun $ $ \Leftarrow $ Shortest distance whitin bin $(l\_b, l\_p) $
		\If{$ min\_dist\_boun < r$}
		\For{each point $p\_temp$ in bin ($l\_b, l\_p$)}
		\If{$dist(p\_temp, q) < r$}
		\State update $list\_rn$ with $p\_temp$;
		\EndIf
		\EndFor
		\State update $list\_search$ with $(l\_b, l\_p)$;
		\EndIf
		\EndFor
		\EndWhile
		\State \Return{$list\_rn$}
	\end{algorithmic}
\end{algorithm}

\section{Experimental results and analysis}\label{sec4}
In this section, the performance of the proposed 2L-LSH method is compared with that of Kd-tree and Octree, in terms of memory occupation, {\it k}NN search time and RN search time. The IDE used in our experiment is Visual Studio 2017, and the code is implemented using C++98. The experiment is conducted on an Intel Core i7-9750H CPU with 16GB RAM.

\subsection{Datasets}\label{sec4_1}
\begin{figure}
	\centering
	\includegraphics[width=3in]{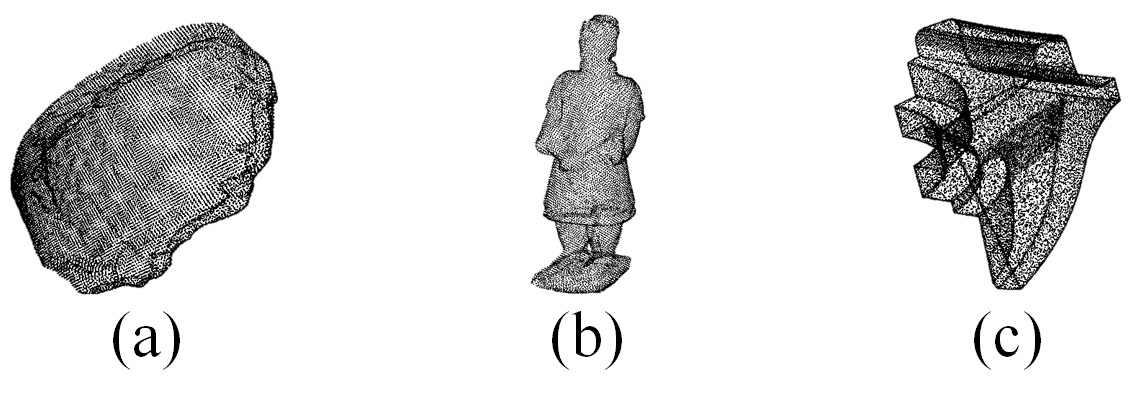}
	\caption{Representative models scanned from real objects. (a) Plate (with 19638points). (b) Terracotta (with 13037points). (c) Disk (with 23937 points).}
	\label{fig6}
\end{figure}

A large number of point cloud models are used for the performance testing to prove the outperformance of the proposed 2L-LSH method. The information of several representative models is presented in Figure~\ref{fig6}. These models are scanned from real objects and have different shapes.

We also introduced open source datasets ModelNet40 \cite{bib25} and ABC \cite{bib26} for experiments to obtain more convincing experimental results. ModelNet40 \cite{bib25} dataset contains Computer-Aided Design (CAD) models from 40 categories, commonly used for training the deep network in 3D deep learning. ABC \cite{bib26} dataset is a collection of CAD models designed for researching  geometric deep learning  methods and applications.

Given that models in practical contexts often have large scales, we exclusively tested models with more than 1,000 points - specifically, 9,294 models in the ModelNet40 \cite{bib25} dataset and 5,073 models in the ABC \cite{bib26} dataset. This approach ensures that our experiments yield more accurate results. The scales of some models in ModelNet40 \cite{bib25} and ABC \cite{bib26} are listed in Table~\ref{tab5}.

\begin{table}
	\begin{center}
		\begin{minipage}{245pt}
			\caption{Some examples in ModelNet40 \cite{bib25} and ABC \cite{bib26} }\label{tab5}%
			\centering
			\begin{tabular}{@{}llll@{}}
				\toprule
				\quad Name of models & \qquad Number of points\\
				\midrule
				\quad bookshelf\_0249 & \qquad 39,550 \\
				\quad tv\_stand\_0126 & \qquad 57,262 \\
				\quad xbox\_0113 & \qquad 34,695 \\
                    \quad abc\_00008473 & \qquad 1,095 \\
				\quad abc\_00004597 & \qquad 5,186 \\
                    \quad abc\_00000893 & \qquad 10,457 \\    
				\bottomrule
			\end{tabular}
		\end{minipage}
	\end{center}
\end{table}

\subsection{Implementation details}\label{sec4_2}
\textbf{Parameters:} The parameter $div$ determines the number of hash bins in the 2L-LSH data structure. Small $div$ will lead to too many points in the hash bin, making it take longer to search the bin. Large $div$ will lead to the need for frequent iterations to expand the search range, which reduces the overall efficiency of the algorithm. As shown in Figure~\ref{fig4}c, after the 2L-LSH hashing process, the OBB space where the target object resides will be divided into $24 \times div$ hash bins. The average number of points divided into each hash bin is denoted by $p_{avg}$, then $m=24 \times div \cdot p_{avg}$. Equation (\ref{equ5}) can be used to calculate $div$ in the 2L-LSH method:
\begin{equation}
	div=\lceil {\frac{m}{24 \cdot p_{avg}}} \rceil
    \label{equ5}
\end{equation}

\begin{figure}
	\centering
	\includegraphics[width=3.3in]{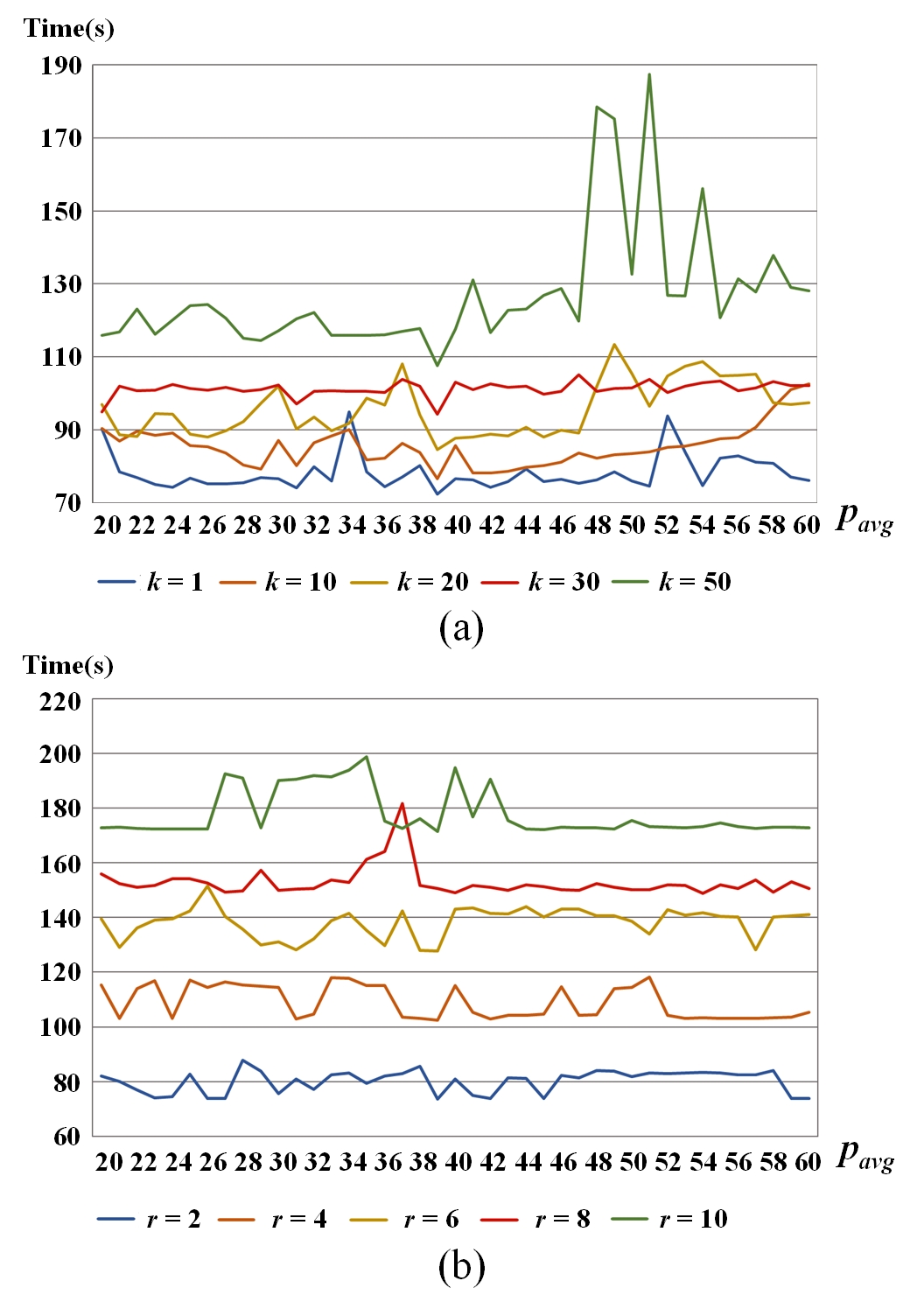}
	\caption{
            Neighboring points search time with different values for $p_{avg}$. (a) {\it k}NN. (b) RN.
            }
	\label{fig7}
\end{figure}

To determine the optimal value of $div$, we randomly selected 1,000 points from each point cloud model to be tested as search points, then perform {\it k}NN and RN searches for each search point and counts the time taken for this work. Taking a model set containing number of $m\in$[5K, 10K) points as an example, the results are shown in Figure~\ref{fig7}, where the horizontal axis represents the value of $p_{avg}$, and the vertical axis represents the time (in seconds) spent completing all \emph{k}NN (RN) searches. As can be observed from the figure, both \textit{k}NN and RN searches demonstrate better performance when the value of $p_{avg}$ is 39.

Furthermore, we also conducted similar experiments on other models. The experiments show that, the most appropriate parameter $p_{avg}$ value can be selected according to Table~\ref{tab1}(1K=1,000), the parameter $div$ in subsequent experiments are set according to this table and Equation (\ref{equ5}).

\begin{table}
	\begin{center}
		\begin{minipage}{235pt}
			\caption{Optimal parameter for model of different sizes}\label{tab1}%
			\centering
			\begin{tabular}{@{}llll@{}}
				\toprule
				\quad Number of points $m$ & \qquad \qquad $p_{avg}$ & \qquad \qquad     $layer$\\
				\midrule
				\quad $m < $5K\qquad & \qquad \qquad 15\qquad & \qquad \qquad 4  \\
				\quad $m \in $[5K, 10K)\qquad & \qquad \qquad 39\qquad & \qquad \qquad 4  \\
				\quad $m \in $[10K, 100K)\qquad & \qquad \qquad 48\qquad & \qquad \qquad 6 \\
				\quad $m \in $[100K, 200K)\qquad & \qquad \qquad 95\qquad & \qquad \qquad 7\\
				\quad $m \geq $200K\qquad & \qquad \qquad 381\qquad & \qquad \qquad 8  \\
				\bottomrule
			\end{tabular}
		\end{minipage}
	\end{center}
\end{table}

\begin{figure}
	\centering
	\includegraphics[width=3.3in]{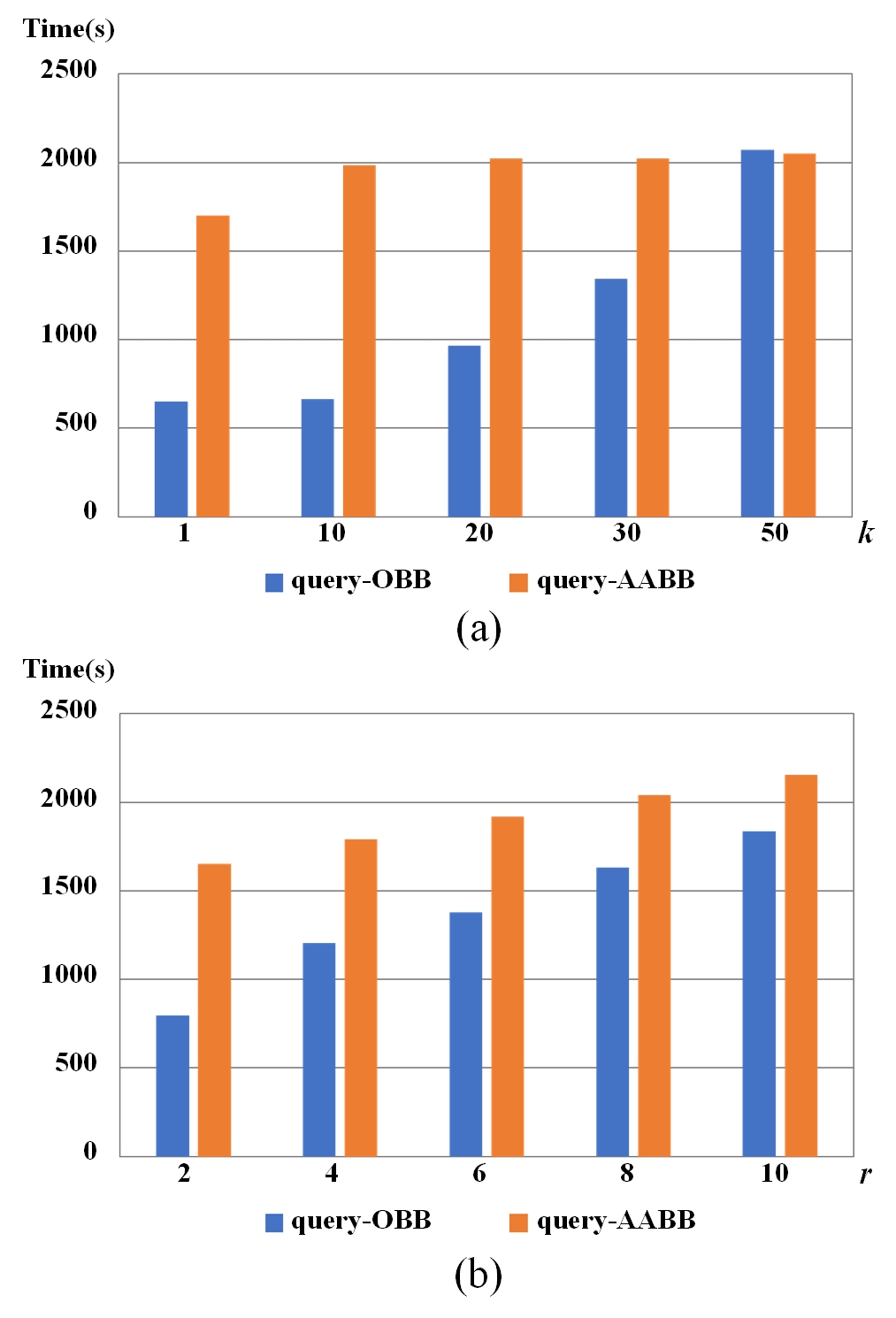}
	\caption{
            Neighboring points search time with different bounding boxes. (a) {\it k}NN. (b) RN.}
	\label{fig10}
\end{figure}

\textbf{The effectiveness of OBB:} The proposed 2L-LSH method is constructed based on OBB instead of AABB because AABB contains more “empty space” than OBB. We also conducted experiments to validate this claim. For each test model in dataset ModelNet40 \cite{bib25} and ABC \cite{bib26}, we selected 1,000 points and performed {\it k}NN and RN search using the 2L-LSH method based on AABB and OBB, respectively. According to the experimental results, the average time consumption for establishing AABB and OBB is 14.309 seconds and 3.128 seconds, respectively. The statical search time consumption are shown in Figure~\ref{fig10}, where the horizontal axis represents the $k$($r$) value and the vertical axis represents the search time (in seconds). From the experimental results, it can be seen that although the time consumption of establishing AABB for the point cloud model is slightly lower than OBB, the 2L-LSH method based on AABB has a much higher time consumption for neighboring points searching than the OBB-based method.

\textbf{Parameters of Octree and Kd-tree: }The proposed algorithm is then compared with the most commonly used Octree \cite{bib5} and Kd-tree \cite{bib6} algorithms, we therefore details the parameters of Octree and Kd-tree. 

Octree requires to generate the minimum bounding box of the point cloud model, and then divide it into small cubes having the same size. This division is iterated until the tree height reaches a specified value. Each small cube  is assigned a unique index value, and all the points in the same cube share the same index value. It can be deduced from the experiments that for models of different sizes, the height of Octree, denoted by parameter $layer$, which can achieve an optimal effect, is also different. The $layer$ value that is too large will result in a vast search space and unnecessary time consumption, while the value that is too small may lead to excessive iterations in the search process and memory overflow. Through experiments, we found the $layer$ values that are most suitable for models of different sizes and list them in Table~\ref{tab1}. The parameters $layer$ in subsequent experiments are set according to this table. 

Kd-tree enables a quick search for neighboring points of the specified search point. More precisely, the tree is traversed to locate the nearest region, which is most likely to contain the neighboring points of the search point, and then this region is traversed. If the target point is close to the region boundary, the backtracking method is required to generate the correct result \cite{bib6}. While this method boasts small memory consumption and fast search speed, making it widely popular in practical use, its challenge lies in the frequent branch judgments during the Kd-tree establishment process, rendering the tree-building process time-consuming.

\subsection{Memory occupation}\label{sec4_2}
Two important criteria to measure the quality of an algorithm are time consumption and memory occupation. It is usually difficult to achieve the optimal value of both at the same time, so it is necessary to balance those two. The tree structure-based methods are usually considered as time saving as well as memory efficient \cite{bib27}. For a point cloud containing $m$ points, the memory occupied by the pointers used to construct the data structure in Octree, Kd-tree and 2L-LSH method 
are $\mathrm{O}\left(\sum_{i=1}^{layer} 8^{i-1}+m\right)$, $\mathrm{O}\left(\sum_{i=1}^{\log_2(m+1)} 2^{i-1}\right)$ and $\mathrm{O}\left(24\times div+m\right)$, respectively. Assuming that a pointer occupies 4 bytes of memory, the memory occupied by the pointers in the data structures of Octree, Kd-tree and 2L-LSH method are presented in Table~\ref{tab2}. It can be seen from the table that although the establishment and use of Octree is relatively simple, the cost is a serious waste of space. Furthermore, the proposed method has roughly the same memory occupation as Kd-tree, which is much lower than that of Octree, especially when dealing with large scale models, for example, when the point cloud contains more than 2,500,000 points, the memory occupied by Octree is almost twice that of 2L-LSH.

\begin{table}
	\begin{center}
		\begin{minipage}{245pt}
			\caption{Memory occupation of pointers}\label{tab2}%
			\begin{tabular*}{\textwidth}{@{\extracolsep{\fill}}lcccccc@{\extracolsep{\fill}}}
				\toprule
				\multirow{2}{*}{}
				Number of points  & 2L-LSH & Kd-tree & Octree\\
				{\it m} & (KB) & (KB) & (KB)\\
				\midrule
				5K & 20.51 & 19.53 & 21.82 \\
				10K & 40.63 & 39.06 & 185.35 \\
				100K & 403.65 & 390.63 & 1560.91 \\
				200K & 790.44 & 781.25 & 1951.54 \\
				2,500K & 9880.51 & 9765.63 & 19127.91 \\
				\bottomrule
			\end{tabular*}
		\end{minipage}
	\end{center}
\end{table}

\subsection{Experimental results}\label{sec4_4}
In this section, we will present, analyze and summarize the experimental results on the datasets introduced in Section~\ref{sec4_1} in terms of two aspects: {\it k}NN search and RN search.

\begin{figure}
	\centering
	\includegraphics[width=3.3in]{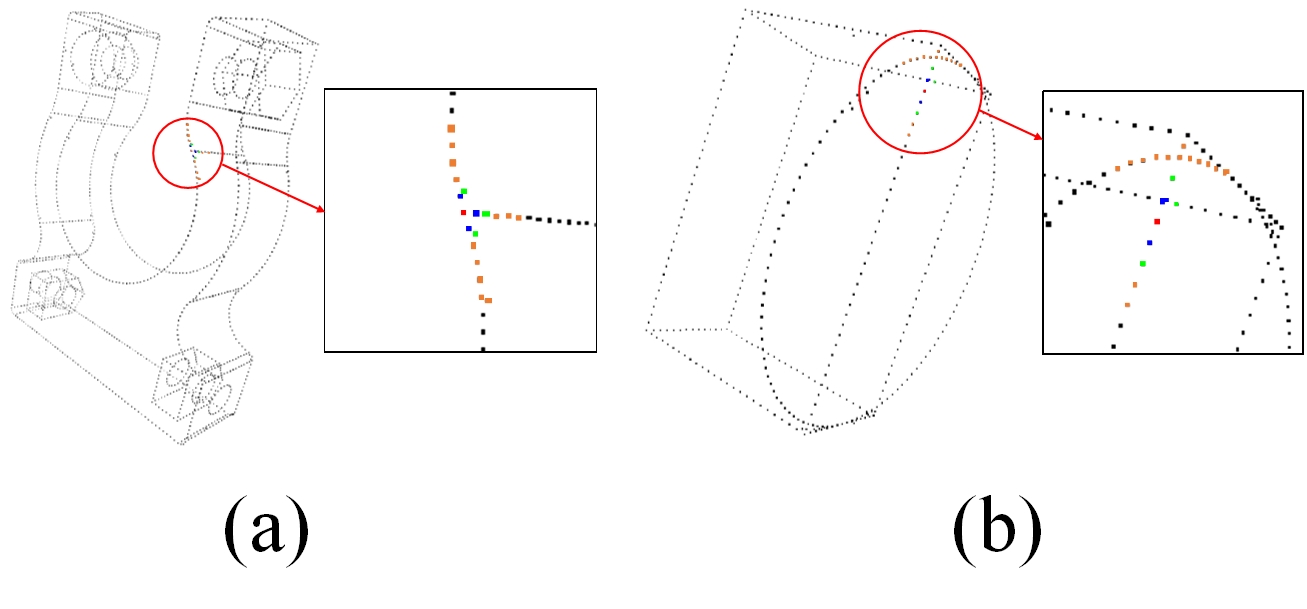}
	\caption{
            The Visualized results. (a) {\it k}NN search. (b) RN search.
            }
	\label{fig11}
\end{figure}

{\it k}NN search is finding the $k$ nearest points based on the Euclidean distance from the search point. And RN search consists in finding all the points within radius $r$ from the search point. The purpose of \textit{k}NN and RN search is to accurately find all the neighboring points of the search point, which means the Overall Ratio, Recall and Precision must be 100\%. To display the results visually, we marked the search point and its 3 and 6 nearest neighboring points in red, blue and green, respectively, with the other neighboring points were marked in orange. The visualized results are shown in Figure~\ref{fig11}, where we set $k$=20 and $r$=3. 

We denote the time consumption of neighboring points searching using 2L-LSH, Kd-tree and Octree by $t_{2LLSH}$, $t_{Kdtree}$ and $t_{Octree}$, respectively. The difference $Reduct\_1$ between the search time of Kd-tree and 2L-LSH, and the difference $Reduct\_2$ between Octree and 2L-LSH can be calculated using Equation(\ref{equ6}) and (\ref{equ7}):
\begin{equation}
	Reduct\_1={\frac{t_{Kdtree}-t_{2LLSH}}{t_{Kdtree}}}\times100\% \label{equ6}
\end{equation}
\begin{equation}
        Reduct\_2={\frac{t_{Octree}-t_{2LLSH}}{t_{Octree}}}\times100\% \label{equ7}
\end{equation}

 \textbf{\textit{k}NN search:} In the experiment, 1,000 points are randomly selected as search points for each point cloud model shown in Figure~\ref{fig6} and Table~\ref{tab5}, and then the $k$ nearest neighbors ($k$=1,2,3,4,5) are calculated for each search point. In Table~\ref{tab3}, the obtained results are presented to visually demonstrate the effectiveness of the proposed 2L-LSH algorithm. It can be seen that, 2L-LSH can be 50.138\% and 94.159\% faster than Kd-tree and Octree. 

\begin{table*}
	\begin{center}
		\begin{minipage}{\textwidth}
			\caption{The search times of {\it k}NN. The best scores are in bold.}\label{tab3}%
			\begin{tabular*}{\textwidth}{@{\extracolsep{\fill}}lcccccc@{\extracolsep{\fill}}}
				\toprule
				\multirow{2}{*}{}
				Model & $k$ & 2L-LSH & Kd-tree & $Reduct\_1$ & Octree & $Reduct\_2$\\
				&  & (ms) & (ms) & (\%) & (ms) & (\%)\\
				\midrule
				\multirow{5}{*}{plate}
				& 1 & \textbf{204} & 363 & 43.802 & 469 & 56.503 \\
				& 2 & \textbf{181} & 363 & 50.138 & 485 & 62.680 \\
				& 3 & \textbf{197} & 352 & 44.034 & 491 & 59.878 \\
				& 4 & \textbf{189} & 352 & 46.307 & 486 & 61.111 \\
				& 5 & \textbf{190} & 351 & 45.869 & 511 & 62.818 \\
				\hline
				\multirow{5}{*}{terracotta}
				& 1 & \textbf{85} & 130 & 34.615 & 280 & 69.643 \\
				& 2 & \textbf{83} & 126 & 34.127 & 299 & 72.241 \\
				& 3 & \textbf{82} & 132 & 37.879 & 313 & 73.802 \\
				& 4 & \textbf{88} & 135 & 34.815 & 299 & 70.569 \\
				& 5 & \textbf{85} & 134 & 36.567 & 306 & 72.222 \\
				\hline
				\multirow{5}{*}{disk}
				& 1 & \textbf{222} & 425 & 47.765 & 554 & 59.928 \\
				& 2 & \textbf{220} & 425 & 48.235 & 589 & 62.649 \\
				& 3 & \textbf{222} & 424 & 47.642 & 615 & 63.902 \\
				& 4 & \textbf{221} & 434 & 49.078 & 589 & 62.479 \\
				& 5 & \textbf{221} & 430 & 48.605 & 611 & 63.830 \\
                    \hline
                    \multirow{5}{*}{bookshelf\_0249}		
                & 1 &  \textbf{78} & 94 & 17.021 & 513 & 84.795 \\
				& 2 & \textbf{87} & 91 & 4.396 & 559 & 84.436 \\
				& 2 & \textbf{75} & 84 & 10.714 & 417 & 82.014 \\
				& 4 & \textbf{67} & 125 & 46.400 & 404 & 83.416 \\
				& 5 & \textbf{72} & 137 & 47.445 & 415 & 82.651 \\
                    \hline
                    \multirow{5}{*}{tv\_stand\_0126}			
                & 1 & \textbf{54} & 97 & 44.330 & 597 & 90.955 \\
				& 2 & \textbf{59} & 87 & 32.184 & 603 & 90.216 \\
				& 3 & \textbf{44} & 69 & 36.232 & 491 & 91.039 \\
				& 4 & \textbf{40} & 79 & 49.367 & 464 & 91.379 \\
				& 5 & \textbf{45} & 81 & 44.444 & 521 & 91.363 \\
				\hline
                    \multirow{5}{*}{xbox\_0113}
                & 1 & \textbf{34} & 43 & 20.930 & 378 & 91.005 \\
				& 2 & \textbf{33} & 46 & 28.261 & 565 & 94.159 \\
				& 3 & \textbf{46} & 56 & 17.857 & 453 & 89.845 \\
				& 4 & \textbf{33} & 43 & 23.256 & 409 & 91.932 \\
				& 5 & \textbf{31} & 41 & 24.390 & 438 & 92.922 \\
                   
                    \hline
                    \multirow{5}{*}{abc\_00008473}
                    & 1 & \textbf{20} & 25 & 20.000 & 71 & 71.831 \\
                    & 2 & \textbf{22} & 45 & 51.111 & 116 & 81.034 \\
                    & 3 & \textbf{20} & 31 & 35.484 & 95 & 78.947 \\
                    & 4 & \textbf{27} & 34 & 20.588 & 87 & 68.966\\
                    & 5 & \textbf{24} & 41 & 41.463 & 117 & 79.487\\
                    \hline
                    \multirow{5}{*}{abc\_00004597}
                    & 1 & \textbf{52} & 73 & 28.767 & 205 & 74.634\\
                    & 2 & \textbf{67} & 78 & 14.103 & 243 & 72.428\\
                    & 3 & \textbf{50} & 63 & 20.635 & 238 & 78.992\\
                    & 4 & 67 & 103 & 34.951 & 254 & 73.622 \\
                    &5 & \textbf{48} & 68 & 29.412 & 211 & 77.251 \\ 	 	
                    \hline
                    \multirow{5}{*}{abc\_00000893}
                    & 1 & \textbf{83} & 161 & 48.447 & 304 & 72.697 \\
                    & 2 & 105 & 169 & 37.870 & 338 & 68.935 \\
                    & 3 & \textbf{127} & 156 & 18.590 & 260 & 51.154 \\
                    & 4 & \textbf{72} & 119 & 39.496 & 233 & 69.099 \\
                    & 5 & \textbf{75} & 126 & 40.476 & 253 & 70.356 \\
                    \bottomrule
			\end{tabular*}
			\footnotetext{$Reduct\_1$ and $Reduct\_2$ are the differences between the {\it k}NN search time of Kd-tree and 2L-LSH, Octree and 2L-LSH, respectively.}
		\end{minipage}
	\end{center}
\end{table*}

\begin{figure}
	\centering
	\includegraphics[width=3.3in]{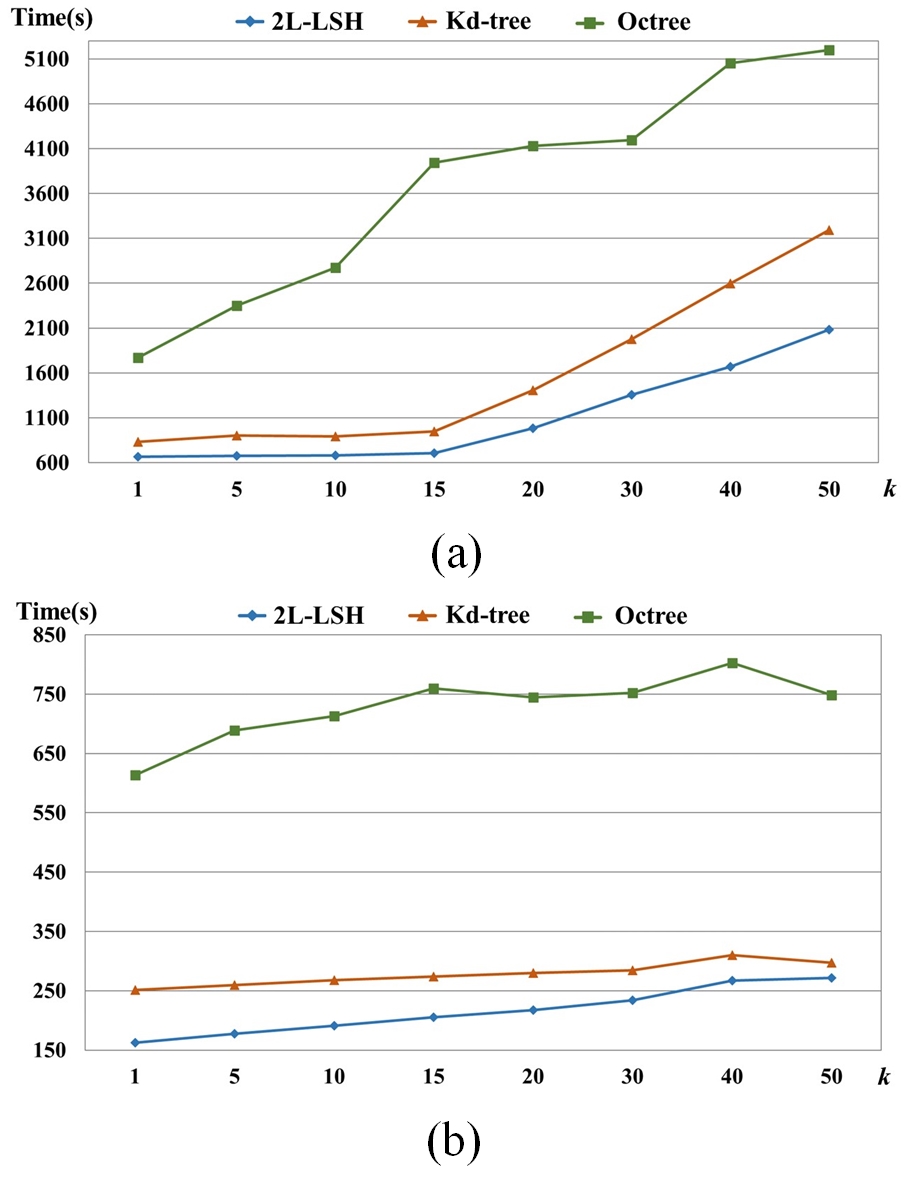}
	\caption{The calculating times of {\it k}NN search. (a) On ModelNet40 \cite{bib25}. (b) On ABC \cite{bib26}.}
	\label{fig8}
\end{figure}

On dataset ModelNet40 \cite{bib25}, our method can also achieve better performance when selecting 1,000 points in each model to operate {\it k}NN search. The results can be seen in Figure~\ref{fig8}a, where the horizontal axis represents the value of $k$ and the vertical axis denotes the time spent in {\it k}NN search (in seconds). When $k$=50, 2L-LSH is 1107.86 seconds faster than Kd-tree and 3116.32 seconds faster than Octree. We also ran the same experiment on ABC dataset \cite{bib26}, and the experimental results are shown in Figure~\ref{fig8}b. When $k$=10, 2L-LSH is 76.728 seconds faster than Kd-tree and 522.411 seconds faster than Octree.

It can be seen that although the time-consuming trend varies slightly on different datasets, in general, among all {\it k}NN searches, the proposed 2L-LSH method is the fastest, the Octree is the slowest, and the speed of Kd-tree is in the middle.

\textbf{RN search:} Similarly, 1,000 points are randomly selected from each model shown in Figure~\ref{fig6} and Table~\ref{tab5}. 

The RN search is operated when $r$=2, 4, 6, 8, 10), and the times of RN search are counted for all the search points. The statistical results are listed in Table~\ref{tab4}. It can be seen that 2L-LSH is the fastest among them, and can be 51.025\% and 39.274\% faster than Kd-tree and Octree.

\begin{table*}
	\begin{center}
		\begin{minipage}{\textwidth}
			\caption{Statical RN search times}\label{tab4}%
			\begin{tabular*}{\textwidth}{@{\extracolsep{\fill}}lcccccc@{\extracolsep{\fill}}}
				\toprule
				$r$ & 2L-LSH(ms) & Kd-tree(ms) & $Reduct\_1$(\%) & Octree(ms) & $Reduct\_2$(\%)\\
				\midrule
                    2 & \textbf{1228} & 2700 & \textbf{54.519} & 1643 & 25.259 \\
                    4 & \textbf{1479} & 2907 &  49.123 & 2543 & \textbf{41.840} \\
                    6 & \textbf{1698} & 3056 &  44.437 & 2712 &  37.389 \\
                    8 &  \textbf{1967} & 3165 & 37.852 & 2772 & 29.040 \\
                    10 & \textbf{2257} & 3291 & 31.419 & 2941 & 23.257\\
				\bottomrule
			\end{tabular*}
			\footnotetext{$Reduct\_1$ and $Reduct\_2$ are the differences between the RN search time of Kd-tree and 2L-LSH, Octree and 2L-LSH, respectively.}
		\end{minipage}
	\end{center}
\end{table*}

\begin{figure}
	\centering
	\includegraphics[width=3.3in]{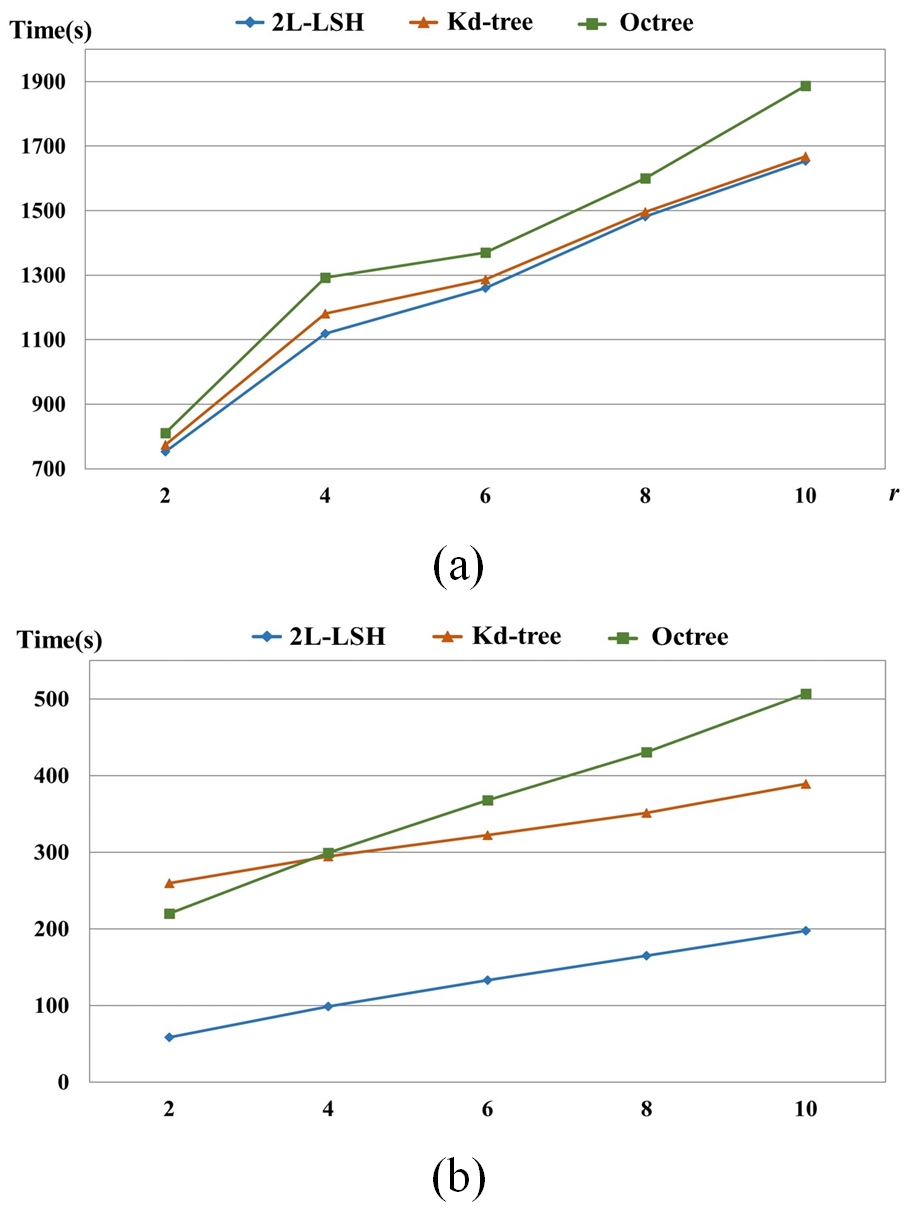}
	\caption{The calculating times of RN search. (a) On ModelNet40 \cite{bib25}. (b) On ABC \cite{bib26}.}
	\label{fig9}
\end{figure}

For ModelNet40 \cite{bib25}, 1,000 points are also selected in each model to test RN search. The results are shown in Figure~\ref{fig9}a, where the horizontal axis and vertical axis represent the value of $r$ and the time spent in RN search (in seconds), respectively. It can be seen that, Kd-tree is faster than Octree, but both are slower than 2L-LSH. When $r$=4, 2L-LSH is 62.164 seconds faster than Kd-tree and 172.886 seconds faster than Octree. 

The same operation as the above experiment was carried out on ABC \cite{bib26} dataset , and the results are shown in Figure~\ref{fig9}b. When $r$=2, the speed of Kd-tree is slightly slower than Octree, and when $r\geq4$, the speed of Kd-tree is faster than Octree. But in general, both Kd-tree and Octree are still slower than the proposed 2L-LSH.

\begin{figure}
	\centering
	\includegraphics[width=3.3in]{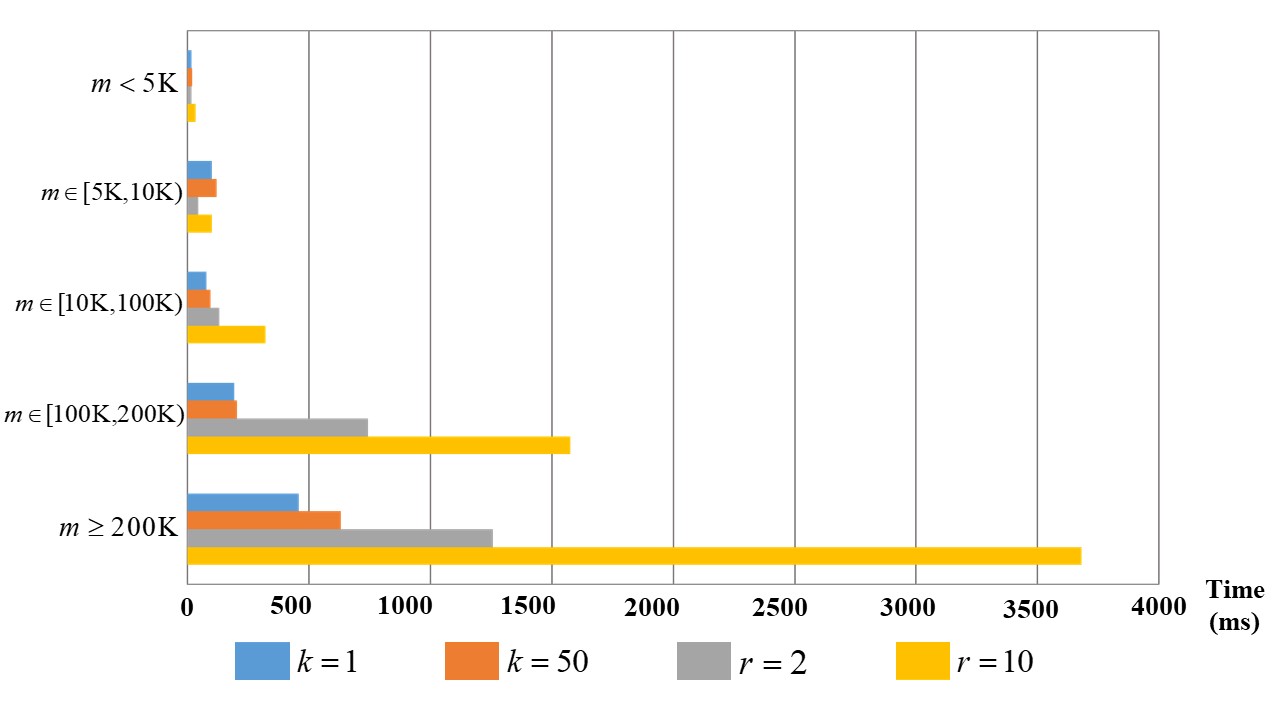}
	\caption{
            Experiment results of point cloud points and computing time.}
	\label{fig12}
\end{figure}

We also conducted experiments on point clouds with different scales of points to analyze the efficiency of 2L-LSH with respect to the size of point clouds. For ModelNet40 \cite{bib25} and ABC \cite{bib26}, 1,000 points are randomly selected in each model to operate {\it k}NN and RN searches. According to the data division in Table 1, we conducted the above experiments on the point clouds whose scales are falling into each interval, and then calculated the average time consumption of the data within that interval. The results are shown in Figure~\ref{fig12}, where the horizontal axis and vertical axis represent the points number $m$ and the average time spent in neighbors points searching (in milliseconds), respectively. It can be seen that, for  point clouds with smaller scale, neighboring points searches based on 2L-LSH can perform well. As the number of points on the point cloud increases, the search speed gradually decreases, and the changes in RN search are more significant than that of {\it k}NN search.

\section{Conclusion}\label{sec5}

In this paper, a method based on LSH, referred to as 2L-LSH, is proposed to structure the point cloud. Using the two-level hash function, the search space of the neighboring points is significantly reduced, thereby greatly decreasing the time cost of {\it k}NN and RN searches.  The proposed  2L-LSH is compared with the the popular Octree and Kd-tree to validate its time and memory efficiency. Although the proposed method achieves a high performance, it still faces the problem of selecting appropriate parameter values. In future work, we will aim at adaptive computing of the parameters to further improve the applicability of the proposed algorithm.

\nocite{*}

\bibliographystyle{compj}
\bibliography{ModellingBidders}

\end{document}